\documentclass{article}

\usepackage[nonatbib,preprint]{neurips_2020}
\usepackage[numbers]{natbib}         

\usepackage[utf8]{inputenc} 
\usepackage[T1]{fontenc}    
\usepackage{hyperref}       
\usepackage{url}            
\usepackage{booktabs}       
\usepackage{amsfonts}       
\usepackage{nicefrac}       
\usepackage{microtype}      

\usepackage{graphicx}       
\usepackage{float}          
\usepackage{multirow}       
\usepackage{multicol}
\usepackage{rotating}

\usepackage{subfloat}
\usepackage[%
    format=hang,    
    format=plain,
    margin=0pt,
    width=0.8\textwidth,
]{caption}
\usepackage{mwe}
\usepackage[list=true]{subcaption}
\usepackage[export]{adjustbox}
\usepackage{forloop}
\usepackage{etoolbox}

\usepackage{amsmath}
\usepackage{listings} 
\usepackage{pgfplots} 
\usepackage{wrapfig, blindtext}

\pgfplotsset{compat=newest}
\usetikzlibrary{shadings,shadows,shapes.geometric,arrows,fit,matrix,
positioning,calc,arrows.meta, decorations.pathmorphing,shapes.symbols}

\pgfplotsset{width=10cm,compat=1.9}

\title{Uncertainty Intervals for Graph-based Spatio-Temporal Traffic Prediction}

\author{%
  Tijs Maas \\
  Informatics Institute, \\
  University of Amsterdam \\
  \texttt{tijs\_maas@hotmail.com} \\
   \And
   Peter Bloem \\
  INDELab, \\
University of Amsterdam\\
  \texttt{uva@peterbloem.nl} \\
}

\begin{document}

\maketitle

\begin{abstract}
Many traffic prediction applications rely on uncertainty estimates instead of the mean prediction. Statistical traffic prediction literature has a complete subfield devoted to uncertainty modelling, but recent deep learning traffic prediction models either lack this feature or make specific assumptions that restrict its practicality. We propose Quantile Graph Wavenet, a Spatio-Temporal neural network that is trained to estimate a density given the measurements of previous timesteps, conditioned on a quantile. Our method of density estimation is fully parameterised by our neural network and does not use a likelihood approximation internally. The quantile loss function is asymmetric and this makes it possible to model skewed densities. This approach produces uncertainty estimates without the need to sample during inference, such as in Monte Carlo Dropout, which makes our method also efficient.  
%
\end{abstract}

\section{Introduction}
Currently, most traffic prediction models predict the average traffic conditions from minutes 5 up to an hour ahead. While impressive, this problem setting largely ignores the uncertainty of the generated predictions, making the results more difficult to interpret. Many real-world applications rely on confidence intervals or certainty bounds for these predictions, instead of the mean predicted value. For example, when scheduling road maintenance or when planning a route with minimal delays.

Using Monte Carlo (MC) Dropout \citep{gal_dropout_2016}, it is possible to model the trips within a city, and utilize vehicle trajectories to predict future traffic speeds \cite{zhu_deep_2017, zhang_boosted_2019}. This technique requires B stochastic forward passes to compute the sample variance. Increasing the number of stochastic inference samples B improves the quality variance. Monte Carlo (MC) Dropout makes the assumption that uncertainty in traffic speed and volume can be modelled as multivariate Gaussian. The reality however, is that these distributions are skewed and asymmetric and thus do not satisfy the assumption made here. For instance, when the traffic speed is measured to be maximum, it is far more likely to decrease than increase.

In this extended abstract, we introduce a novel traffic prediction model based on the spatio-temporal neural network Graph Wavenet \cite{wu_graph_2019}. However, instead of optimizing our mean prediction, we train using a Quantile loss function, similar to Autoregressive Implicit Quantile Networks \cite{ostrovski_autoregressive_2018}. This makes it possible to estimate the density at specific points, conditioned on a quantile. Our method allows for non-gaussian uncertainty modelling, which remains greatly unexplored in the deep learning traffic prediction literature. The contribution of this work can be summarized as follows:

%

\begin{itemize}
\item We introduce a method to approximate asymmetric and skewed density functions to model the uncertainty estimate.

\item Instead of approximating the variance using multiple forward passes, our technique requires only one pass for each requested density quantile. 
\end{itemize}

\section{Quantile Regression}

Quantile Regression is a method to estimate the quantile function of a distribution at chosen points, which is equal to the inverse cumulative distribution function (cdf). 
It has been shown that when minimized using stochastic approximation, quantile regression converges to the true quantile function value \cite{koenker_quantile_2001}. This allows us to approximate a distribution using a neural network approximation of its quantile funtion, acting as reparameterization of a random sample from the uniform distribution.

Let us define the quantile regression loss $\rho_\tau(u) = (\tau - \mathbb{I}[u \leq 0]) u$ \citep{koenker_quantile_2001} for the error $u$ and the quantile $\tau\in \mathcal{U}[0,1]$. When $u$ is positive $F$ underestimates $Z$ i.e. the estimate falls short of the true value. Now for a given scalar distribution $Z$ with cdf $F_Z$ and a quantile $\tau$ we obtain the inverse cdf $F^{-1}_Z(\tau) = q$, which minimizes the expected quantile regression loss $\mathbb{E}_{z\sim Z}[\rho_\tau(z - q)]$.
\begin{figure}[H]
    \centering
\includegraphics[width=0.5\textwidth]{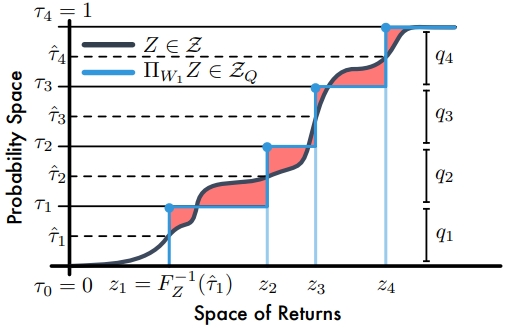}%
\caption{$1$-Wasserstein minimizing projection $\prod_W$ onto N = 4
uniformly weighted Diracs. Shaded regions sum to form the
$1$-Wasserstein error. For detailed explanation, we refer the reader to \citep{dabney_distributional_2017}.}
\label{fig:uncertaintypred3}
\end{figure}

\section{Graph WaveNet}
Graph WaveNet is neural network architecture for spatio-temporal traffic prediction. Different from the AIRAI competition, this model assumes traffic measurements in the form of sensors on a road network graph. The architecture consists of temporal causal convolutions (TCN) \citep{dauphin_language_2017} with graph diffusion convolutions applied to every layer. 

\subsection{Temporal Convolutions}
Gating mechanisms are critical in recurrent neural networks and they have been shown to be powerful to control information flow through layers for temporal convolution networks as well. The gating mechanism in Graph Wavenet is two parallel TCN layers configured with a gate:
\begin{equation}
\mathbf{h}=g\left(\boldsymbol{\Theta}_{1} \star \mathcal{X}+\mathbf{b}\right) \odot \sigma\left(\boldsymbol{\Theta}_{2} \star \mathcal{X}+\mathbf{c}\right)
\end{equation}
Specifically, the LSTM-style gating $(g=\text{tanh})$ also shared with PixelCNN \citep{oord_conditional_2016} and WaveNet \citep{oord_wavenet_2016} is used.
\begin{figure}[H]
    \centering
    \hfill
    \includegraphics[width=0.3\textwidth]{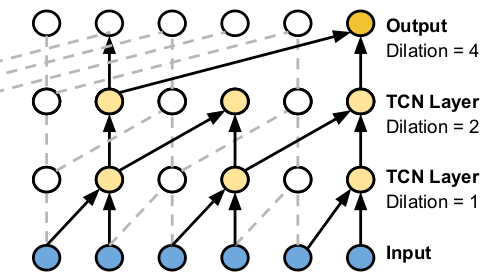}
    \hfill
 \captionsetup{width=.9\linewidth}
    \hfill
\caption{Dilated casual convolution with kernel size 2 and dilation factor k, it picks inputs every k step and applies the standard
1D convolution to the selected inputs.}
\label{fig:infill_methods}
\end{figure}

Another advantage is that the temporal receptive field grows exponentially w.r.t. the number of layers and the dilation factor. To achieve this, we
artificially design the receptive field size of Graph WaveNet
equals to the sequence length of the inputs so that in the last
spatial-temporal layer the temporal dimension of the outputs
exactly equals to one. After that we set the number of output
channels of the last layer as a factor of step length T to get
our desired output dimension.

\subsection{Spatial Graph Diffusion Convolution}
Graph Wavenet uses spatial graph convolution \citep{li_diffusion_2018} to share information on the graph structure. This type of graph convolution differs from the popularised GCN of Kipf and Welling \cite{kipf_semisupervised_2017}, since it is based on epidemic diffusion \cite{lerman_network_2012}:
\begin{align}
\boldsymbol{X}_{:, c} \star_{\mathcal{G}} f_{\boldsymbol{\theta}}=\sum_{k=0}^{K-1}\left(\theta_{k, 1}\left(\boldsymbol{D}_{O}^{-1} \boldsymbol{W}\right)^{k}+\theta_{k, 2}\left(\boldsymbol{D}_{I}^{-1} \boldsymbol{W}^{\top}\right)^{k}\right) \boldsymbol{X}_{:, c} \quad \text { for } c \in\{1, \cdots, C\}
\end{align}
where $\theta\in\mathbb{R}^{N\times 2}$ are the filter parameters and $\boldsymbol{D}_{O}^{-1} \boldsymbol{W}, \boldsymbol{D}_{I}^{-1} \boldsymbol{W}^T$ represent the transition matrices of the diffusion process and the reverse one, respectively. They note that since $T_{k+1} = (D^{-1}_o W) T_k(x)$ and 
$D^{-1}_o W$ is sparse, it is possible to use recursive sparse-dense matrix multiplication arriving at a time complexity $O(K|\mathcal{E}|)$ of their update function, which is similar in complexity to the method proposed by Kipf and Welling. However, unlike the method of Kipf and Welling, the edges between the nodes are non-symmetric. Another advantage is that this method only uses a sparse graph neighbourhood in its update function, rather than the full graph Laplacian, and is therefore more resilient against structural changes.

\subsection{Loss Function}
The training objective is the Mean Absolute Error (MAE) over Q prediction timesteps, N locations, each with C different measurements. 
\begin{equation}
L\left(\hat{\mathbf{X}}^{(t+1):(t+Q)} ; \mathbf{\Theta}\right)=\frac{1}{Q N C} \sum_{i=1}^{i=Q} \sum_{j=1}^{j=N} \sum_{k=1}^{k=C}\left|\hat{\mathbf{X}}_{j k}^{(t+i)}-\mathbf{X}_{j k}^{(t+i)}\right|
\end{equation}

\section{Quantile Graph WaveNet}
We modify Graph WaveNet to implicitly predict the cdf, instead of optimizing for the mean prediction error. This involves conditioning the input on a quantile $\tau$ and training with a quantile loss function.

One drawback to the original quantile regression loss is that gradients do not scale with the magnitude of the error, but instead with the sign of the error and the quantile weight $\tau$ \cite{ostrovski_autoregressive_2018}. The Huber quantile regression loss introduces a threshold $\kappa$, such that if the error is within the threshold $\kappa$, scaling is performed w.r.t. the magnitude of the errors. In our experiments we find that $\kappa=0.05$ works well.
\begin{align*}
\rho^\kappa_\tau(u) &= |\tau - \mathbb{I}\{u \leq 0\}| \mathcal{L}_\tau^\kappa (u),
\text{ with \ }
\mathcal{L}_\tau^\kappa (u) = 
\begin{cases}
\frac{1}{2\kappa} u^2 & \text{if } |u| < \kappa \\
 |u| - \frac{1}{2}\kappa & \text{otherwise}\\
\end{cases}
\end{align*}

\section{Data Preparation}
We evaluate our model on traffic in Los Angelos (METR-LA \citep{jagadish_big_2014}), Istanbul and Berlin\citep{kreil_surprising_2020}. The dataset METR-LA recorded 207 loop detectors in the metropolitan area of Los Angelos from March 1st, 2012 until June 30th, 2012. We partition this timeline into 3 non-overlapped sections: training, validation and test with the respective ratios 7:1:2. 

The Traffic4cast competition offers more complicated real-world datasets from the cities: Berlin, Istanbul and Moscow. The data is presented as grid with the resolution of 495x436 pixels for each city, and every pixel consitutes 100m$^2$. The training and validation set contain 181 and 18 days, respectively. In their main challenge it is expected from participants to make 500 predictions of up to 1h into the future (test set), spread over 163 days.

The Berlin and Istanbul dataset are an excellent opportunity to test our uncertainty prediction model on a larger scale. However, Graph WaveNet expects a sensors on a graph, which is different from the 495x436x9 image in the competition. In theory, one could consider every pixel as a node on the graph, however it is the case that most pixels report traffic measurements infrequently. In contrast to METR-LA, where the percentage of operational sensors is typically $>90\%$, as seen in Table \ref{tbl:1}. For practicality reasons, we sampled pixels from the 495x436x9 image with a value density of at least $\frac{1}{16}$ in the outskirts and $\frac{1}{2}$ towards the city centre. From the remaining pixels we sample equidistant points (d = 1200m) on the traffic graph, in Berlin this yields a graph of approximately 1300 nodes. Measurements of the average traffic speed are used at an interval of 5 minutes.

\begin{figure}[H]
  \centering
      \includegraphics[width=0.2\textwidth]{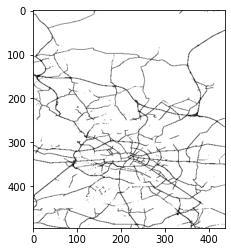}        
          \parbox[c]{0.1\textwidth}{\Large$\Rightarrow$\\\vspace{4.5em}}
\includegraphics[width=0.2\textwidth]{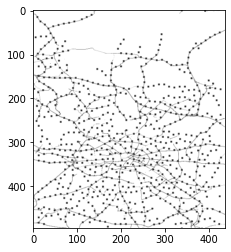}\vspace{-4.5em}\caption{Extraction of equidistant measurement locations on the traffic network.}
\end{figure}

\begin{table}[]
\centering
\begin{tabular}{l|l|l|l}
\hline 
$\mathcal{D}$ & Measurement locations & Duration & No measurement\\  \hline
METR-LA & 207 & 34272 & 0.08 \\
ISTANBUL & 999 & 52128 & 0.61 \\ 
BERLIN & 1336 & 52128 & 0.77 \\ 
\hline
\end{tabular}
\vspace{1em}
\caption{Traffic dataset statistics (after preprocessing).}
\label{tbl:1}
\end{table}

%
%

\section{Results}
We evaluate the results of our model on the datasets METR-LA, ISTANBUL and BERLIN. Since we do not have the complete test set of ISTANBUL and BERLIN, we instead have partitioned the 181 days training set into train, validation and test with the ratio: (0.89, 0.01, 0.10). Our main results are the uncertainty estimates and their calibration, we compare this to MC Dropout uncertainty estimates. Additionally, we also provide the mean prediction accuracy and compare this with previous methods.

\begin{figure}[H]
  \centering
      \includegraphics[width=0.27\textwidth]{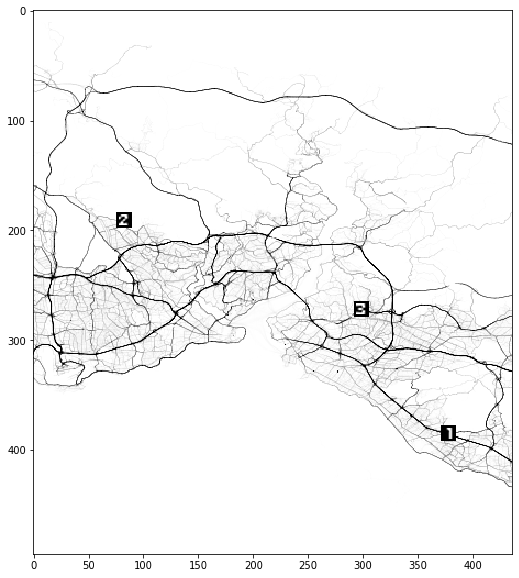}
      \includegraphics[width=0.615\textwidth]{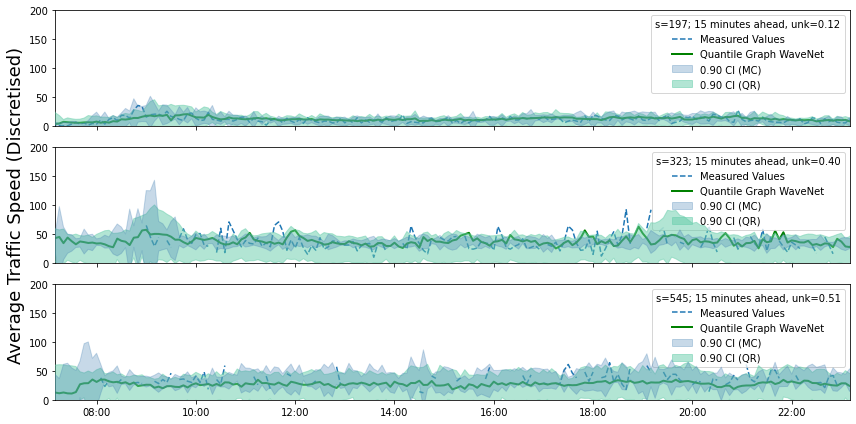}
  \caption{Randomly selected sensors grouped by percentage of given measurements.}
  \label{fig:3}
\end{figure}

\textbf{Qualitative results} To demonstrate that we can successfully learn uncertainty intervals that are skewed and asymmetric, we plot the 0.9 CI in Figure \ref{fig:3}, for both Graph WaveNet with the proposed Quantile estimation and Graph WaveNet with MC Dropout. We also project the mean prediction of the original measurements. The locations of the sensors are visible in the map on the left.

A benefit of the Quantile uncertainty interval (lightgreen) is that each boundary or mean is generated by a distinct $\tau$ value, on which the network is conditioned. This is efficient to compute and greatly improves the flexibility of uncertainty intervals, which is expressed as skewed and asymmetric densities.

\begin{figure}[H]
  \centering
      \includegraphics[width=0.3\textwidth]{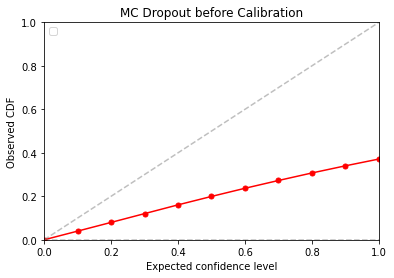}
      \includegraphics[width=0.3\textwidth]{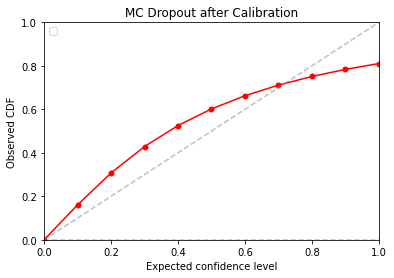}
      
      \includegraphics[width=0.3\textwidth]{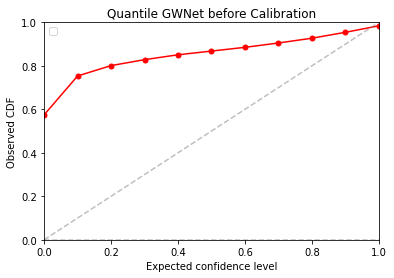}
      \includegraphics[width=0.3\textwidth]{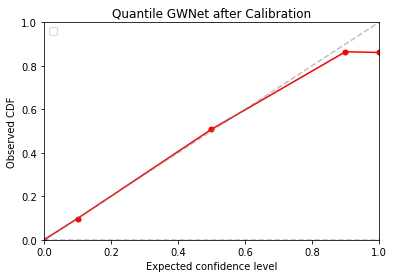}
  \caption{Expected confidence vs observed cdf, before and after calibration on the validation set.}
  \label{fig:2}
\end{figure}

\textbf{Calibrated uncertainty}
After training, we need to calibrate our model on the validation set. Figure \ref{fig:2} shows the uncertainty calibration of BERLIN: before calibration, after calibration on the validation set and on the test set. 

Both Monte Carlo Dropout and Quantile Graph WaveNet require an additional calibration step, however the advantage of the latter is that we can re-map every tau value, as suggested by \cite{kuleshov_accurate_2018}. The underestimation that is present in many Bayesian Neural networks \cite{kuleshov_accurate_2018} is not present in the Quantile Graph WaveNet. However, recalibration by reassigning the Quantiles is not perfect either since exploring the region outside of what is learned may cause the density to decrease rather than increase as wished. We do assume the CDF to start with 0.0 and end with 1.0, but there is no mathematical law that requires this start- and endpoint. In theory we can shrink and expand our space if needed, but retraining on these areas with remapped tau values may help prevent reversal of the density.

\textbf{Mean prediction accuracy}
Lastly, we compare the mean prediction accuracies for the default Graph Wavenet and our proposed Quantile Graph Wavenet (denoted by Quantile est in table below). Note that this is just to demonstrate the difference and this is not the main contribution of this research.

\begin{adjustbox}{center}
\bgroup
\def\arraystretch{1.1}
\scriptsize\fontsize{10}{12}\selectfont
\begin{tabular}[b]{p{0.2cm} p{2.7cm} 
 p{1cm}  p{1cm}| p{1cm} p{1cm}| p{1cm}  p{1cm}}
\hline 
  \multirow{2}{*}{$\mathcal{D}$}&\multirow{2}{*}{Model}& \multicolumn{2}{c}{15 min}& \multicolumn{2}{c}{30 min} & \multicolumn{2}{c}{60 min}\\
  &  & 
   MAE & MSE & MAE & MSE & MAE & MSE \\ \hline
  \multicolumn{1}{l|}{\multirow{5}{*}{
  \raisebox{-.4in}{\rotatebox{90}{METR-LA}}
\renewcommand\arraystretch{1.2}}}
& Hist. Average & 5.18 & 81.18 &  5.18 & 81.18   & 5.18 & 81.18  \\
\multicolumn{1}{l|}{} 
& Static Pred. & 4.03 & 75.86 & 5.11 & 124.32  & 6.82 & 202.77   \\
\cline{2-8} 
\multicolumn{1}{l|}{} 
& Graph WaveNet$^\dagger$ & 2.69 & 26.62 & 3.08 & 38.81  & 3.50 & 53.14   \\
\multicolumn{1}{l|}{} 
& $\hookrightarrow$ Dropout MC$^\dagger$ & 2.71 & 27.14  & 3.13 & 39.81  & 3.60 & 55.50   \\
\multicolumn{1}{l|}{} 
& $\hookrightarrow$ Quantile est.$^\dagger$ & 3.90 & 38.68  & 3.95 & 42.25 & 4.05 & 45.56 \\
 
 \hline 
  \multicolumn{1}{l|}{\multirow{5}{*}{
  \raisebox{-.1in}{\rotatebox{90}{ISTANBUL}}
\renewcommand\arraystretch{1.2}}}
& Hist. Average* & 23.63 & 852.64 & 23.63 & 852.64 &  23.63 & 852.64 \\
\multicolumn{1}{l|}{} 
& Static Pred. & 11.06 & 313.99 &   11.22 & 320.05 & 11.53 & 331.96 \\
\cline{2-8}
\multicolumn{1}{l|}{} 
& Graph WaveNet$^\ddag$ & 5.58  & 131.79    & 5.66   & 136.30 & 5.80  & 144.00 \\
\multicolumn{1}{l|}{} 
& $\hookrightarrow$ Dropout MC$^\ddag$ & 5.80 & 129.53 &  5.89 & 134.00  & 6.09 & 141.14   \\
\multicolumn{1}{l|}{} 
& $\hookrightarrow$ Quantile est.$^\ddag$ & 6.16 & 168.66 &   6.32 & 176.82   & 6.61 & 192.22 \\\hline 

 \hline 
  \multicolumn{1}{l|}{\multirow{5}{*}{
  \raisebox{.1in}{\rotatebox{90}{BERLIN}}
\renewcommand\arraystretch{1.2}}}
& Hist. Average* & 19.18 & 520.29 & 19.18 & 520.29 &  19.18 & 520.29 \\
\multicolumn{1}{l|}{} 
& Static Pred. & 12.48 & 778.96 &   12.59 & 791.85 & 12.77 & 807.69  \\
\cline{2-8}
\multicolumn{1}{l|}{} 
& Graph WaveNet$^\ddag$ & 7.38  & 234.09    & 7.45   & 244.60& 7.61  & 252.81 \\
\multicolumn{1}{l|}{} 
& $\hookrightarrow$ Quantile est.$^\ddag$ & 7.74 & 318.00 &  7.83  & 326.05  & 8.00 & 338.48\\ 
& \hspace{1em} $\hookrightarrow$ after calibr.$^\ddag$ & 7.77 & 320.41 &  7.86  & 328.38  & 8.02 & 340.76\\
\hline
\end{tabular}
\egroup
\end{adjustbox}
$\dagger:$ The mean of 3 experiments. \\
$*:$ Historical weekly averages, predictions of missing values are counted as average target prediction error.\\
$\ddag:$ Results of 1 experiment, the adaptive adjacency matrix is not enabled in the model. \\

We observe a decrease in the mean prediction accuracy if we compare Graph Wavenet with its Quantile estimation counterpart, and this is consistent for all datasets. A possible explanation why Quantile Graph WaveNet performs worse on the Traffic4cast datasets may be due to more frequently missing measurements. The versatility of Quantile Graph WaveNet can also be a weakness, using the same number of parameters to predict for any quantile, almost certainly reduces the complexity of the mean approximation. For instance if the density to be learned is shaped differently from the mean.
Quantile Graph Wavenet predicts on a subset of the pixelspace used by the Traffic4cast competition and direct accuracy score comparison with other contestents should be avoided. However, making a fair comparison is possible by masking the pixels not used in the graph and computing the MSE of the active pixels. This table omits many recently proposed models and for a comprehensive study on traffic prediction models, we refer the reader to \citep{yin_comprehensive_2020}. 

\section{Discussion}
The method presented here is a step towards reliable traffic prediction uncertainty estimates. Initial results appear promising but we also want to highlight the limitations of our method.
\begin{enumerate}
\item Crossing quantiles: define two quantiles $\tau_1$ and $\tau_2$, with $\tau_1 < \tau_2$, then we should never have that $f(x, \tau_1) < f(x, \tau_2)$.
\item Uncertainty missing for missing values: estimating a density when there are no values has no solution.
\item Additional calibration step is required after training.
\end{enumerate}
Since uncertainty intervals are highly useful in traffic prediction applications, we believe that densities beyond the Gaussian should be considered the topic of future research. We chose to model the density directly, following a technique developed in Reinforcement Learning, thereby omitting the likelihood computation all together. In conjunction, more descriptive confidence metrics such as \citep{laves_wellcalibrated_2019} and \citep{levi_evaluating_2020} could be developed with traffic prediction uncertainty applications in mind.

\section*{Broader Impact}
Improvements in traffic prediction have the potential to improve traffic conditions and reduce travel delays by facilitating better utilization of the available capacity. Traffic uncertainty estimation allows better planning for scheduled road maintanance and uncertainty estimates are more informative when routing critical trips.

\begin{ack}
%
%
Work done partly during intership at HAL24K.
\end{ack}

\bibliographystyle{unsrtnat}
\bibliography{neurips_2020}

\end{document}